%% file: AnonymousSubmission2027.tex
\documentclass[letterpaper]{article}
\usepackage[preprint]{aaai2027}
\usepackage[hyphens]{url}
\usepackage{graphicx}
\usepackage{natbib}
\usepackage{caption}
\usepackage{float}
\usepackage{multirow}

\usepackage{booktabs}
\usepackage{amsmath}
\usepackage{amssymb}
\usepackage{pifont}

\title{LaST$_{0.5}$: A Generalist World Action Model with Interaction-Centric \\Latent Reasoning}
\author{
    Xuheng Zhang\textsuperscript{1,*},
    Yiming Zhang\textsuperscript{1,*},
    Jialin Gao\textsuperscript{2,*},
    Hao Chen\textsuperscript{2,*,\textdagger},
    Jiaming Liu\textsuperscript{1,*,\textdagger},
    Ji Li\textsuperscript{1,*},\\
    Runzhong Huang\textsuperscript{1},
    Yinxi Wang\textsuperscript{1},
    Chenyang Gu\textsuperscript{1},
    Zihan Yang\textsuperscript{1},
    Yaowei Liu\textsuperscript{1},
    Zhonghao Yan\textsuperscript{1},\\
    Ziyu Guo\textsuperscript{2},
    Renrui Zhang\textsuperscript{2},
    Peng Jia\textsuperscript{3},
    Pheng-Ann Heng\textsuperscript{2},
    Shanghang Zhang\textsuperscript{1,\ding{41}}
}
\affiliations{
    \textsuperscript{1}State Key Laboratory of Multimedia Information Processing,
    School of Computer Science, Peking University\\
    \textsuperscript{2}The Chinese University of Hong Kong
    \quad \textsuperscript{3}Simplexity Robotics\\
    \textsuperscript{*}Equal contribution
    \quad \textsuperscript{\textdagger}Project leads
    \quad \textsuperscript{\ding{41}}Corresponding author
}

\begin{document}

\maketitle

\input{Sections/abstract}
\input{Sections/Introduction}
\input{Sections/RelatedWork}
\input{Sections/Method}
\input{Sections/Experiment}
\input{Sections/Conclusion}
\bibliography{aaai2027}

\clearpage
\appendix
\section*{Appendix}
\input{Sections/appendix}

\end{document}

%% file: Sections/abstract.tex
\begin{abstract}

World Action Models (WAMs) have recently emerged as a promising paradigm for robotic manipulation, endowing policies with strong capabilities for modeling temporal and physical dynamics. Existing WAMs, encompassing both video-first action prediction and joint video–action modeling, directly transfer global physical representations from world models to action generation. However, global scene representations may underemphasize local robot–object interactions, creating a mismatch between representations optimized for world evolution and those required for manipulation.
To bridge this gap, we propose LaST$_{0.5}$, a novel WAM that introduces an interaction-centric physical representation between world modeling and action generation. Specifically, LaST$_{0.5}$ learns a compact interaction-centric latent space under the supervision of visual interaction cues between the robot and the environment. This reasoning space is instantiated as a latent expert that bridges the world expert and the action expert within a Mixture-of-Transformers architecture. Furthermore, LaST$_{0.5}$ performs adaptive world-to-action transfer by jointly predicting future representations and progress values, using progress consistency to adjust action execution and trigger future resampling.
Finally, we pretrain LaST$_{0.5}$ on a large-scale robotic data corpus comprising over 14k hours of data, establishing interaction-centric physical reasoning as a scalable and effective paradigm. LaST$_{0.5}$ achieves 99.5\% on LIBERO, surpasses the previous state of the art on ten evaluated RLBench tasks, and demonstrates robust long-horizon execution and precise dexterous manipulation in real-world experiments.

\end{abstract}

%% file: Sections/Introduction.tex
\section{Introduction}

\begin{figure*}[t]
    \centering
    \includegraphics[width=\textwidth]{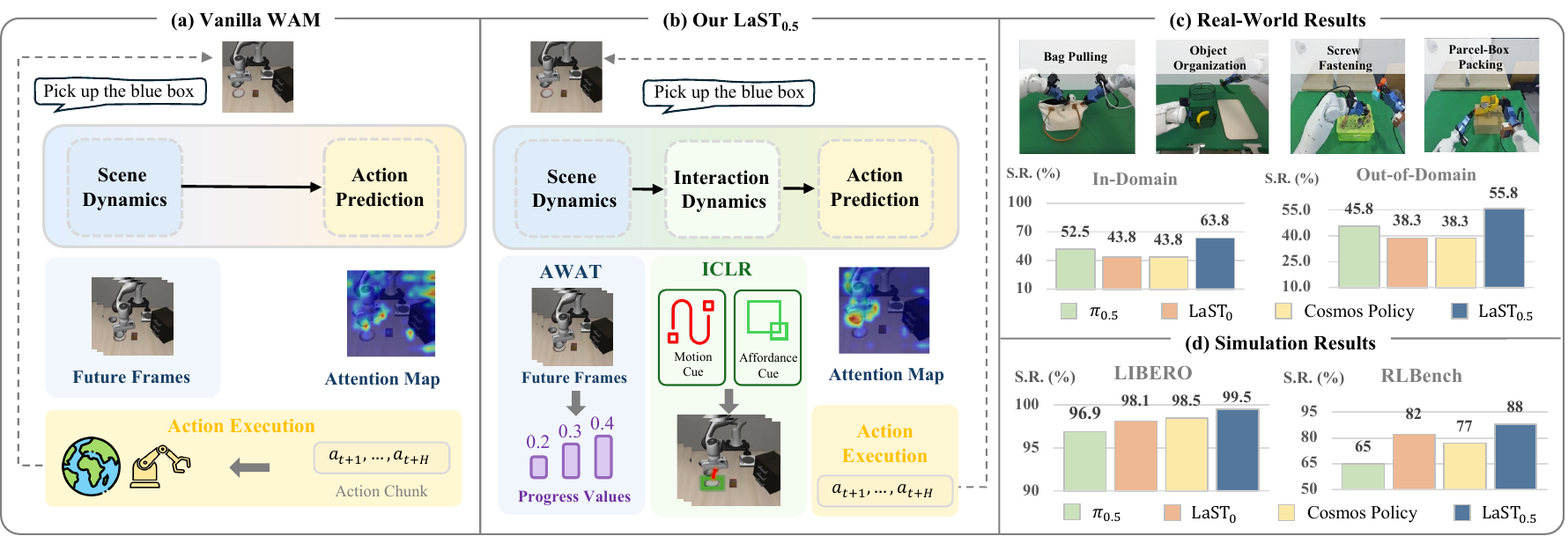}
    \vspace{-0.4cm}
    \caption{\textbf{Overview.} Existing WAMs directly transfer representations of global scene dynamics to action prediction, potentially overlooking the localized robot–environment interactions required for manipulation. LaST$_{0.5}$ bridges this gap through Interaction-Centric Latent Reasoning (ICLR), which learns a compact latent space supervised by visual motion and object-affordance cues, establishing a hierarchical reasoning paradigm that progressively transforms global world representations into interaction-centric abstractions and ultimately into robot actions. LaST$_{0.5}$ achieves state-of-the-art performance on the LIBERO and RLBench simulation benchmarks and demonstrates robust long-horizon and precise manipulation in real-world tasks.
    }
    \label{fig:teaser}
    \vspace{-0.3cm}
\end{figure*}

Building general-purpose robotic foundation models requires policies not only to understand the current scene but also to reason about how the physical world evolves under robot actions.
While Vision-Language-Action (VLA)~\cite{kim2024openvla,kim2025finetuningvisionlanguageactionmodelsoptimizing,liu2025hybridvla,intelligence2025pi05visionlanguageactionmodelopenworld,2025_6_02_Fast_in_Slow,gu2025manualvla,qu2025spatialvla} models have achieved strong performance by directly mapping visual observations to actions, World Action Models (WAMs)~\cite{yuan2026fast,ye2026world,ye2026gigaworld,bi2026motus,ye2026world} further augment this paradigm with world modeling, enabling policies to leverage temporal causality, implicit physics, and motion priors learned from video data.
By jointly modeling world evolution and robot actions, WAMs provide a more generalizable and transferable paradigm for robotic learning~\cite{kim2026cosmos}.
Despite these advantages, effectively translating world knowledge into robot actions remains an open challenge.

Existing WAMs can be broadly categorized into two paradigms: \emph{Imagine-then-Execute}, which first predicts future observations and then infers actions through inverse dynamics~\cite{li2026causal,du2023learning,hu2024video,feng2026harmowam}, and \emph{Joint Modeling}, which jointly predicts future observations and robot actions~\cite{bi2026motus,ye2026world,yuan2026fast,luo2026being}. As shown in Figure~\ref{fig:teaser}, both paradigms share a common assumption: representations optimized for modeling world evolution inherently facilitate action generation.
Although world modeling provides rich physical knowledge for manipulation, its global prediction objective may underemphasize the fine-grained robot–object interactions required for precise manipulation. Consequently, directly transferring global world representations may not fully exploit the physical knowledge required for action generation, motivating an interaction-centric physical abstraction that bridges world modeling and robot actions.

To bridge this gap, we propose \textbf{LaST$_{0.5}$}, a generalist WAM built upon an \textbf{Interaction-Centric Latent Reasoning (ICLR)} framework, as shown in Figure~\ref{fig:teaser}. Unlike existing WAMs that directly transfer global world representations to action generation, LaST$_{0.5}$ first learns a compact latent space that transforms global world representations into interaction-relevant physical abstractions. This reasoning space is instantiated as a latent expert within a Mixture-of-Transformers architecture~\cite{liang2024mixture}, receiving global representations from a pretrained video world expert~\cite{ali2025world}. To construct interaction-centric latent supervision, future frames are augmented with interaction visual cues encoding manipulation motion and target-object affordance, and encoded by a native vision encoder into latent targets. These latent targets encourage the latent expert to learn compact physical representations centered on robot-environment interactions. Conditioned on this latent space, the efficient action expert generates precise robot actions. Overall, LaST$_{0.5}$ follows a hierarchical reasoning paradigm that progressively transforms global world representations into interaction-centric abstractions and ultimately into robot actions.
Furthermore, we introduce an \textbf{Adaptive World-to-Action Transfer (AWAT)} mechanism that enables the world model to jointly predict future representations and their corresponding progress values. Since fluctuations in world modeling accuracy can directly compromise action generation, AWAT identifies future frames with increasing progress values and executes only the temporally aligned actions from the corresponding action chunks, using progress consistency to dynamically determine the execution horizon.

To empirically validate our framework, we first pretrain LaST$_{0.5}$ on a large-scale heterogeneous robotic data corpus comprising over 14k hours of data \cite{wu2024robomind,o2024open,khazatsky2024droid}, establishing interaction-centric physical reasoning as a scalable and effective paradigm. As shown in Figure~\ref{fig:teaser}, LaST$_{0.5}$ achieves an average success rate of 99.5\% on the LIBERO benchmark and an 88.0\% success rate on RLBench, outperforming the previous state-of-the-art (SOTA) methods. In real-world deployments, we validate the effectiveness of the interaction-centric latent reasoning in LaST$_{0.5}$ across four bimanual tasks using the Tianji robot.
LaST$_{0.5}$ demonstrates robust long-horizon execution and precise control on challenging manipulation tasks. It also exhibits zero-shot generalization to unseen objects, backgrounds, and object positions, further validating the capability and robustness of our unified world-to-interaction-to-action modeling design.
In summary, our main contributions are as follows:

\begin{itemize}

    \item We bridge the gap between global scene modeling in world models and the interaction-centric reasoning required for action generation by introducing LaST$_{0.5}$, a generalist WAM with an Interaction-Centric Latent Reasoning (ICLR) framework.

    \item We introduce the Adaptive World-to-Action Transfer (AWAT) mechanism, which jointly predicts future representations and their corresponding progress values to dynamically assess the reliability of world knowledge, to adapt action execution and trigger future resampling based on progress consistency.

    \item We pretrain LaST$_{0.5}$ on a massive corpus of over 14k hours of robotic data, achieving SOTA results on simulation benchmarks (99.5\% on LIBERO and 88.0\% on  the ten evaluated RLBench tasks) and demonstrating exceptional long-horizon execution and precise manipulation in real-world tasks.

\end{itemize}

%% file: Sections/RelatedWork.tex
\section{Related Work}

\subsection{World Action Models for Robotic Manipulation}

World Action Models (WAMs) have recently emerged as a promising paradigm for robotic manipulation by leveraging temporal dynamics and physical priors learned from large-scale video generation models. Existing approaches mainly follow imagine-then-execute pipelines that ground predicted visual evolution into executable actions~\cite{du2023learning,bharadhwaj2024gen2act,jang2025dreamgenunlockinggeneralizationrobot,feng2026harmowam}, or joint video-action modeling that predicts future observations and robot actions within a unified framework~\cite{bi2026motus,ye2026world,kim2026cosmos,li2026causal}. Researchers further improve inference efficiency by reducing explicit pixel-space prediction~\cite{yuan2026fast,luo2026being}. However, typical WAMs transfer representations optimized for global scene evolution directly to action generation, overlooking the precise interaction dynamics. In contrast, LaST$_{0.5}$ introduces interaction-centric latent reasoning to distill global world knowledge into compact physical abstractions tailored for manipulation.

\subsection{Visual Prompting for Robotic Manipulation}

Recent studies have explored visual prompts as explicit interfaces of structured manipulation guidance beyond implicit visual understanding. Existing approaches introduce visual cues including points, sketches, and spatial annotations to specify task-relevant interaction regions, enabling more precise robot control~\cite{liu2024moka,yuan2024robopoint,sundaresan2024rt,gu2025manualvla}. Beyond static spatial cues, subsequent works incorporate lines, arrows, and motion trajectories to represent manipulation intentions and temporal interaction procedures~\cite{li2025crayonrobo,tan2026action}. Existing visual prompts primarily provide input-level task instruction, without explicitly modeling the interaction representations for action generation; however, LaST$_{0.5}$ leverages future interaction cues as latent supervision to establish an interaction-centric reasoning space between world modeling and action generation.

\subsection{Value Learning for Robotic Manipulation}

Value learning in robotic manipulation has evolved from task-specific return estimation to transferable task-progress modeling. Traditional approaches typically train visual Q-functions to estimate task-specific returns from interaction data~\cite{kalashnikov2018qtopt,chebotar2023qtransformer}. Beyond it, recent methods learn dense progress signals from vision-language priors, action-free videos, and temporal ordering, enabling more transferable value representations across tasks~\cite{ma2022vip,ma2023liv,ma2024gvl,tan2025robodopamine}. Furthermore, value estimators have been integrated with foundation policies and world models to rank predictions and filter demonstrations~\cite{ren2026vvlaps,kim2026cosmos,li2026worldvalueaction,wang2026worldvalue}. However, previous methods mainly use value as an external evaluation signal over candidate actions, trajectories, or demonstrations. Instead, LaST$_{0.5}$ derives temporally aligned progress from forward model representations and adaptively modulates world-to-action transfer through progress consistency.

%% file: Sections/Method.tex
\section{Method}

\begin{figure*}[t]
    \centering
    \includegraphics[width=0.99\textwidth]{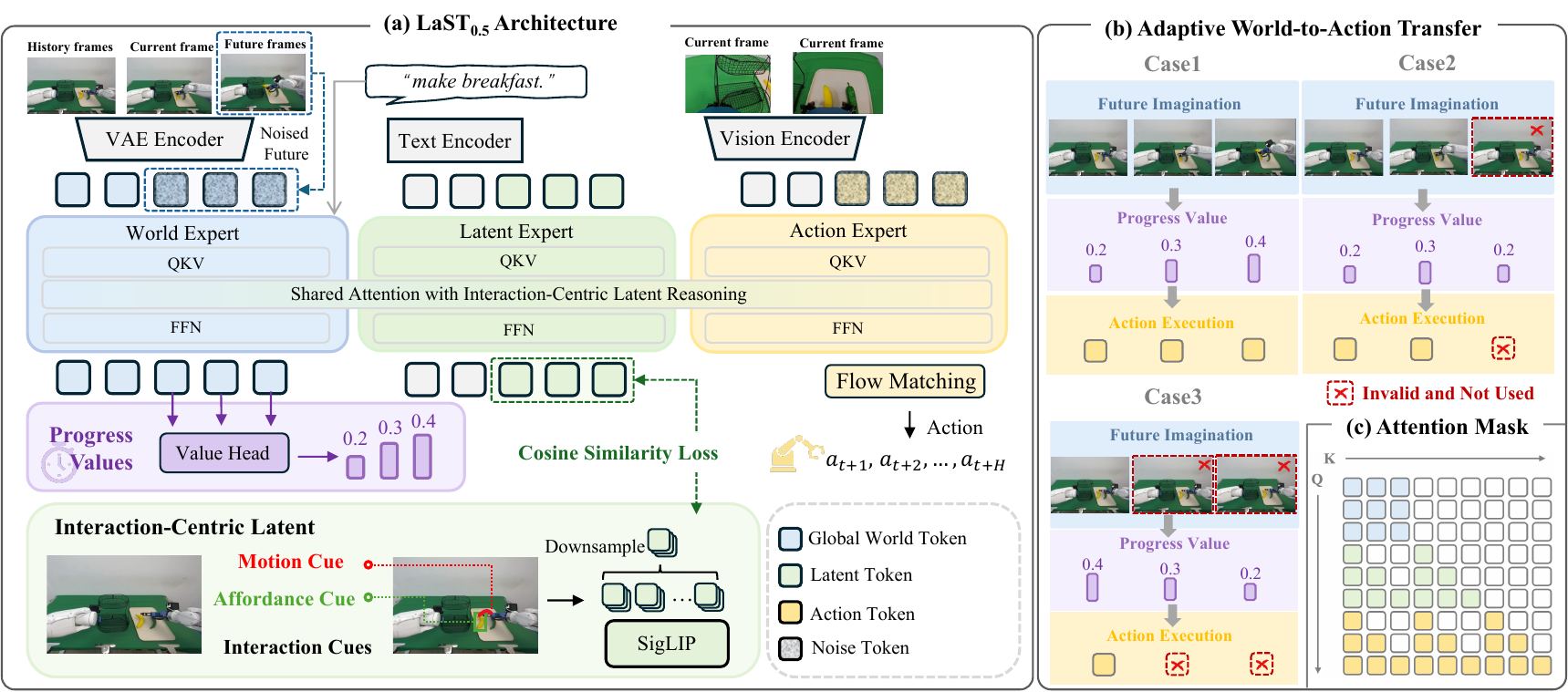}
    \vspace{-0.1cm}
    \caption{\textbf{The LaST$_{0.5}$ Framework.} (a) LaST$_{0.5}$ unifies a World Expert, Latent Expert, and Action Expert within a Mixture-of-Transformers (MoT) architecture. To bridge global scene dynamics and precise local control, the Latent Expert distills future frames into compact, interaction-centric representations. (b) The Adaptive World-to-Action Transfer (AWAT) mechanism predicts progress values to assess task-progress consistency in imagined futures. Action execution is dynamically truncated when progress value drops, adapting the execution horizon to the predicted progress sequence. (c) Attention mask of LaST$_{0.5}$.}
    \vspace{-0.3cm}
    \label{fig:method}
\end{figure*}

\subsection{Problem Formulation}
We consider \emph{vision-language robotic manipulation}: given a natural-language instruction and visual observations of the scene, a robot must execute a sequence of actions to accomplish the specified task. We train a policy $\pi_\theta$ via imitation learning on a large-scale, action-labeled heterogeneous corpus $\mathcal{D}$. This dataset encompasses diverse collection modalities, including teleoperated robot demonstrations, in-the-wild data collected via portable handheld grippers (e.g., UMI~\citep{umi}), and human egocentric data. At each timestep $t$, the policy takes a natural-language instruction $l$ and multi-view visual observations $I_t=\{I_t^k\}_{k=1}^{K}$, predicting an action chunk $\mathbf{a}_{t:t+H}$ over a horizon $H$. Each arm’s action comprises a relative 6-DoF end-effector pose and an embodiment-specific actuator command, such as a binary gripper state or a multi-finger hand configuration.
Detailed specifications for all evaluated embodiments are deferred to the Appendix C.

\subsection{LaST$_{0.5}$ Model Design}
As shown in Figure~\ref{fig:method} (a), LaST$_{0.5}$ realizes the policy $\pi_\theta$ as a unified WAM that jointly reasons over three core quantities: future scene evolution, interaction-centric latent dynamics, and robot actions.
Let $c_t=(I_t,l)$ denote the current observation-instruction context.
We partition the predicted future video, interaction-centric latent sequence,
and action chunk into $M$ temporally aligned segments (e.g., $M=3$ for an action chunk of length 12), denoted by
$\hat{V}=\{\hat{V}_i\}_{i=1}^{M}$,
$\mathcal{Z}=\{z_i\}_{i=1}^{M}$, and
$\mathbf{a}=\{\mathbf{a}_i\}_{i=1}^{M}$, respectively.
We factorize their joint predictive distribution as
\begin{equation}
\begin{aligned}
p(\mathbf{a},\mathcal{Z},\hat{V}\mid c_t)
&= p(\hat{V}\mid c_t)\,
p(\mathcal{Z}\mid\hat{V},c_t)\\
&\quad\cdot p(\mathbf{a}\mid\mathcal{Z},\hat{V},c_t).
\end{aligned}
\label{eq:factorization}
\end{equation}
where $\hat{V}_i$, $z_i$, and $\mathbf{a}_i$ denote the temporally aligned world, latent, and action segments, respectively. As shown in Figure~\ref{fig:method}, our attention mask restricts direct cross-expert attention to corresponding and preceding segments.
Each factor in this distribution is parameterized by a dedicated expert: a \emph{world expert} that predicts global scene evolution, a \emph{latent expert} that infers $\mathcal{Z}$---a compact physical abstraction of upcoming action-relevant interaction dynamics---and an \emph{action expert} that generates the action chunk. These three experts operate within a unified attention stream of a Mixture-of-Transformers (MoT) architecture, explicitly structuring the inference as a principled, hierarchical \emph{video $\rightarrow$ interaction $\rightarrow$ action} reasoning process.

\subsubsection{World Expert.}
As shown in Figure~\ref{fig:method}, the world expert forecasts future evolution conditioned on historical and current observations, as well as the language, extracting a foundational physical prior. We instantiate it with Cosmos-Predict2.5-2B~\citep{ali2025world}, a pretrained diffusion transformer. At timestep $t$, a history of five primary-view frames is compressed by a causal video tokenizer into latent contexts, while the instruction $l$ is efficiently injected via cross-attention using cached text embeddings.
The World Expert processes video frames at $256 \times 256$, while images fed to the VLM branch are separately resized to $224 \times 224$. Conditioned on this context, the world expert predicts the future scene video $\hat{V}$ in Eq.~\ref{eq:factorization} by progressively denoising twelve future frames, which are partitioned into $M$ temporally aligned video segments.
We retain the original Cosmos generation objective for the world expert, denoted as $\mathcal{L}_{\mathrm{world}}$.

\subsubsection{Latent Expert.}
While future scene modeling provides valuable physical priors, explicitly capturing fine-grained robot–environment interactions more directly supports reliable action generation.
To bridge this gap, we introduce a latent expert initialized from Janus~\cite{wu2025janus}, a unified multimodal understanding and generation model. Specifically, conditioned on the future scene representations predicted by the world expert, a small set of learnable query tokens is fed into the latent expert to extract a compact, interaction-centric representation $\mathcal{Z}$. To ensure computational efficiency and holistic reasoning, these latent tokens are generated in a single parallel forward pass.

\subsubsection{Action Expert.} The action expert generates the final action chunk $\mathbf{a}_{t:t+H}$.
Initialized from the same language backbone as the latent expert but maintaining independent parameters, it jointly refines the entire action chunk using a flow-matching objective $\mathcal{L}_{\mathrm{act}}$.
To support heterogeneous embodiments, the action expert uses two dedicated noise tokens and corresponding MLP heads at each timestep to predict wrist motions and actuator commands, respectively. Both outputs adopt the maximum dimensionality across embodiments, with unavailable dimensions zero-padded.

\subsubsection{Mixture-of-Transformers Architecture.}
The three experts operate within a unified MoT architecture, sharing a global attention stream while maintaining independent expert weights. Crucially, we enforce the hierarchical video $\rightarrow$ latent $\rightarrow$ action reasoning defined in Eq.~\ref{eq:factorization} through a \emph{blockwise-causal} attention mechanism.
Further details of the attention mask are provided in Appendix A.2.
We detach the world expert's keys and values to block backward gradient flow from the latent and action experts, treating the predicted scene dynamics as a read-only context to protect the pretrained physical prior.

\subsection{Interaction-Centric Latent Reasoning}
\label{sec:iclr}
Although the world expert captures valuable global scene dynamics, its predictions are heavily dominated by broad environmental features that easily overshadow critical, fine-grained interaction cues. Therefore, we introduce the latent expert to explicitly distill the predicted future into a compact, interaction-centric abstraction. By conditioning the action expert on \emph{both} the global scene and this latent abstraction, we establish a progressive reasoning process that ensures the policy simultaneously benefits from a foundational understanding of the overall environment and a sharp focus on the specific interaction dynamics required for execution.

\subsubsection{Interaction Cues and Latent Target Construction.}

To obtain the interaction-related latent targets, we first partition the twelve future frames into $M$ temporally aligned segments. For each segment, we select its last frame and augment it with two visual interaction cues: a 2D curve representing the corresponding end-effector motion and a bounding box grounding the target-object affordance. The 2D curve is obtained by projecting the demonstrated 3D end-effector trajectory from the world coordinate system onto the image plane using the camera parameters. Grounding DINO~\cite{liu2024grounding} generates candidate object boxes conditioned on the task instruction. We select the interaction target by spatially matching these candidates to the projected tool center point (TCP) at gripper opening/closing events, and associate the selected object with the preceding motion segment. This annotation procedure is performed offline during training-data preparation and is not required at inference. The resulting pixel-space prompts highlight interaction-relevant regions without requiring architectural modifications. As shown in Figure~\ref{fig:method}, each prompted frame is then encoded by a frozen SigLIP~\cite{zhai2023sigmoid} vision tower into $N_{p}$ patch tokens of dimension $d_{v}$. A trainable MLP aggregates the patch features into a single $1\times d_{v}$ representation. After projection and normalization, we obtain $M$ latent reasoning tokens:
$\mathcal{Z}^{\mathrm{GT}}
=
\left\{z_i^{\mathrm{GT}}\right\}_{i=1}^{M}.
$
This visual-prompt mechanism guides the latent-target vision encoder to suppress redundant background information and emphasize patches containing the injected interaction cues.
For latent reasoning supervision, the latent expert is trained to align its predictions with the constructed targets using a cosine similarity objective:
$
\mathcal{L}_{\mathrm{lat}}=1-\frac{1}{M}\sum_{i=1}^{M}\cos\big(\hat{z}_i,\,z^{\mathrm{GT}}_i\big),
$
where $\hat{z}_i$ denotes the $i$-th latent token produced by the expert.
During training, the latent expert is prompted by $M$ learnable query embeddings, which are optimized and retained after training. At inference, these learned queries are directly fed into the latent expert to generate the $M$ interaction-centric latent tokens.
The joint training objective is defined as
$
\mathcal{L}
=
\lambda_{1}\mathcal{L}_{\mathrm{world}}
+
\lambda_{2}\mathcal{L}_{\mathrm{lat}}
+
\lambda_{3}\mathcal{L}_{\mathrm{act}}
$, with coefficients $\lambda_{1}$, $\lambda_{2}$, and $\lambda_{3}$ used to balance the numerical scales of the losses.

\subsection{Adaptive World-to-Action Transfer}
\label{AWAT}

While the world expert provides rich physical priors, its predicted future is not uniformly reliable throughout the entire horizon and may contain hallucinated or physically inconsistent outcomes. For example, in LIBERO, the world model may generate a failed grasp in which the target bowl remains on the table.
However, conditioning action generation on erroneous future scene predictions can lead to severe error accumulation in closed-loop control. We therefore introduce Adaptive World-to-Action Transfer (AWAT), which estimates the progress consistency of the predicted future and uses this estimate to dynamically determine the executable action horizon, as shown in Figure~\ref{fig:method} (b).

\subsubsection{Progress-aware value estimation.}
After joint training, we freeze the base model and attach a lightweight MLP value head, $g_{\phi}$, on the denoised future-frame features produced by the world expert to determine which temporally aligned action segments should be executed.
Consistent with the temporal segmentation defined above, each segment contains four future frames, which are compressed by the Cosmos video tokenizer into video latents. We predict one progress value for each segment. Let $i\in\{1,\ldots,M\}$ index the future segments, and let $\mathcal{H}_i$ denote the set of world model DiT hidden tokens associated with the $i$-th video latent. The predicted progress value is obtained by pooling these tokens:
\begin{equation}
\hat{v}_{i}
=
g_{\phi}\left(\operatorname{Pool}\left(\mathcal{H}_{i}\right)\right).
\end{equation}
To define the supervision target, let $t_i$ denote the median environment timestep of the four frames compressed into the $i$-th video latent. Given a full episode of length $T$, the ground-truth progress target is computed as $v^{\mathrm{GT}}_{i}=\frac{t_i}{T}$. The value head is then optimized using a regression objective over all $M$ future segments:
\begin{equation}
\mathcal{L}_{\mathrm{value}}
=
\frac{1}{M}
\sum_{i=1}^{M}
\ell_{\mathrm{reg}}
\left(
\hat{v}_{i},
v^{\mathrm{GT}}_{i}
\right).
\end{equation}
However, progress values in expert demonstrations are predominantly monotonic, while model-generated trajectories may stagnate or regress during deployment. To expose the value estimator to such failure patterns, we augment a subset of training clips by temporally reversing frames or repeating selected frames, with corresponding progress labels adjusted accordingly—reversed clips yield decreasing progress, and repeated frames yield constant targets. This augmentation ensures that the value estimator remains robust to non-monotonic behaviors at inference time, enabling it to reliably distinguish successful forward progression from stagnation and regression.

\subsubsection{Adaptive action execution.}
By strictly aligning the predicted future video, interaction-centric latent sequence, and action chunk along the temporal dimension, we restrict each group of action tokens to access only the world-expert and dynamic-latent KV states associated with its corresponding frame segment and preceding frame segments.
Consequently, if the predicted scene dynamics degrade in the later frames, this mask blocks direct attention from earlier actions to these world and latent segments. Exploiting this alignment, AWAT dynamically truncates execution based on the predicted value sequence $\{\hat{v}_{1},\ldots,\hat{v}_{M}\}$. The robot executes the action chunk only while the predicted progress values increase, i.e., $\hat{v}_{i} > \hat{v}_{i-1}$.
Once the value drops, execution halts, and a new visual observation is acquired from the robot’s cameras for fresh model inference, thereby preserving reliable early actions and discarding those conditioned on deteriorating predictions.
If the initial predicted value is already lower than the progress estimate from the previous cycle, the entire prediction is deemed unreliable. To recover, we sample multiple world-model futures in parallel. The candidate with the highest value score is selected, providing its KV cache to the latent and action experts to ensure robust generation.
Because training tightly couples actions with future dynamics, predicted progress consistency provides a criterion for selecting the executable action horizon. This allows AWAT to dynamically balance long-horizon foresight with frequent closed-loop re-inference, effectively truncating execution when predicted progress stalls or regresses and triggering future resampling when needed.
In the simulation experiments, we report results for both LaST$_{0.5}$ with AWAT and its base variant without AWAT to isolate the contribution of the adaptive execution scheme.

%% file: Sections/Experiment.tex
\begin{table*}[t]
\centering
\caption{
Comparison of LaST$_{0.5}$ and baselines on RLBench, reporting the mean success rate (\%) across three independent runs.
}
\vspace{-0.2cm}
\label{tab:rlbench_main}
\resizebox{\textwidth}{!}{
\begin{tabular}{lccccccccccc}
\toprule
\textbf{Method}
& \textbf{\shortstack{Close\\Box}}
& \textbf{\shortstack{Close\\Laptop}}
& \textbf{\shortstack{Toilet Seat\\Down}}
& \textbf{\shortstack{Sweep to\\Dustpan}}
& \textbf{\shortstack{Close\\Fridge}}
& \textbf{\shortstack{Phone on\\Base}}
& \textbf{\shortstack{Umbrella\\Out}}
& \textbf{\shortstack{Frame Off\\Hanger}}
& \textbf{\shortstack{Wine at\\Rack}}
& \textbf{\shortstack{Water\\Plants}}
& \textbf{\shortstack{Mean S.R. $\uparrow$\\(mean $\pm$ std.)}} \\
\midrule
OpenVLA-OFT~\cite{kim2025finetuningvisionlanguageactionmodelsoptimizing} & \textbf{100.0} & 65.0 & 60.0 & 30.0 & 80.0 & 30.0 & 30.0 & 20.0 & 20.0 & 15.0 & 45.0 $\pm$ 3.0 \\
$\pi_{0.5}$~\cite{intelligence2025pi05visionlanguageactionmodelopenworld} & 90.0 & 95.0 & 85.0 & 75.0 & \textbf{100.0} & 5.0 & 10.0 & 80.0 & 75.0 & 35.0 & 65.0 $\pm$ 4.0 \\
HybridVLA~\cite{liu2025hybridvla} & 85.0 & 95.0 & \textbf{100.0} & \textbf{90.0} & \textbf{100.0} & 50.0 & 50.0 & 70.0 & 50.0 & 50.0 & 74.0 $\pm$ 4.0 \\
Cosmos Policy~\cite{kim2026cosmos} & 95.0 & 95.0 & \textbf{100.0} & 55.0 & 75.0 & \textbf{80.0} & 85.0 & 75.0 & 55.0 & 55.0 & 77.0 $\pm$ 2.0\\
Fast-WAM~\cite{yuan2026fast} & 95.0 & 95.0 & 95.0 & 75.0 & \textbf{100.0} & 25.0 & 75.0 & 60.0 & 0.0 & 10.0 & 63.0 $\pm$ 3.0 \\
LaST$_{0}$~\cite{liu2026last}  & 95.0 & 95.0 & \textbf{100.0} & 80.0 & 85.0 & 75.0 & 75.0 & 70.0 & \textbf{85.0} & 60.0 & 82.0 $\pm$ 3.0\\
\midrule
LaST$_{0.5}$ w/o AWAT  & \textbf{100.0} & 95.0 & 95.0 & 75.0 & \textbf{100.0} & 75.0 & 85.0 & 80.0 & 80.0 & \textbf{65.0} & 85.0 $\pm$ 2.0 \\
\textbf{LaST$_{0.5}$} & \textbf{100.0} & \textbf{100.0} & 95.0 & 80.0 & \textbf{100.0} & \textbf{80.0} & \textbf{90.0} & \textbf{90.0} & 80.0 & \textbf{65.0} & \textbf{88.0} $\pm$ 3.0 \\
\bottomrule
\end{tabular}
}
\vspace{-0.4cm}
\end{table*}

\section{Experiment}

To evaluate our framework, we first curate a large-scale dataset comprising over 14k hours of heterogeneous robot demonstrations (details in Appendix B.1). This massive corpus supplies the rich, cross-domain physical priors essential for our interaction-centric latent reasoning.

\subsection{Simulation Experiment}

\paragraph{Benchmarks.}
We evaluate LaST$_{0.5}$ on two robotic manipulation benchmarks: RLBench~\cite{james2019rlbenchrobotlearningbenchmark} and LIBERO~\cite{liu2023liberobenchmarkingknowledgetransfer}.
For \textbf{RLBench}, we select $10$ diverse and challenging manipulation tasks executed on a simulated Franka Panda arm within the CoppeliaSim environment, including \textit{Close Box}, \textit{Close Laptop}, \textit{Toilet Seat Down}, \textit{Sweep to Dustpan}, \textit{Close Fridge}, \textit{Phone on Base}, \textit{Take Umbrella Out}, \textit{Frame Off Hanger}, \textit{Wine at Rack}, and \textit{Water Plants}. Observation inputs consist of a single front-view RGB image (resized to $224 \times 224$ for the VLM branch) and the simulator-provided language instruction. Following prior frame-sampling protocols~\cite{ze20243ddiffusionpolicygeneralizable, qin2024anyteleopgeneralvisionbaseddexterous}, we collect $100$ demonstration trajectories per task.
For \textbf{LIBERO}, we evaluate our method on four official task suites—\textit{LIBERO-Spatial}, \textit{LIBERO-Object}, \textit{LIBERO-Goal}, and \textit{LIBERO-Long}—utilizing the officially released training datasets and incorporating both agent-view and wrist-view images for the VLM branch at a $224 \times 224$ resolution.

\paragraph{Baselines.}
We compare against available reported or fine-tuned baselines per benchmark:
\textbf{(1) Vision-Language-Action Models (VLAs):} OpenVLA-OFT~\cite{kim2025finetuningvisionlanguageactionmodelsoptimizing}, $\pi_{0.5}$~\cite{intelligence2025pi05visionlanguageactionmodelopenworld}, HybridVLA~\cite{liu2025hybridvla}, InternVLA~\cite{cai2026internvlaa1unifyingunderstandinggeneration}, and LaST$_{0}$~\cite{liu2026last};
\textbf{(2) World Action Models (WAMs):} Cosmos Policy~\cite{kim2026cosmos}, Motus~\cite{bi2026motus}, and Fast-WAM~\cite{yuan2026fast}.
For all baselines, we initialize from their officially released pretrained checkpoints and finetune them in full compliance with the recommended task-specific hyperparameter configurations from their papers.

\paragraph{Implementation and Evaluation Details.}
All models are trained across tasks in a multi-task setting on an $8 \times$ NVIDIA H20 GPU cluster using the AdamW optimizer~\cite{loshchilov2019decoupledweightdecayregularization}.
We train our model for 100 epochs on LIBERO and 300 epochs on RLBench. On RLBench, we conduct three independent evaluation runs
with different random seeds, with 20 rollouts per task in
each run and 60 rollouts per task in total. On LIBERO, we
report the best success rate obtained during training.

\paragraph{Quantitative Results on RLBench and LIBERO.}
As shown in Tables~\ref{tab:rlbench_main} and~\ref{tab:libero_main}, our full framework, LaST$_{0.5}$, establishes new state-of-the-art results across both benchmarks. On LIBERO, it achieves a 99.5\% average success rate, with near-perfect performance across the \textit{Spatial} (99.7\%), \textit{Object} (100.0\%), \textit{Goal} (99.1\%), and \textit{Long} (99.3\%) suites. On the ten challenging RLBench tasks, LaST$_{0.5}$ reaches a peak mean success rate of 88.0\%, significantly outperforming strong baselines such as OpenVLA-OFT (45.0\%), $\pi_{0.5}$ (65.0\%), HybridVLA (74.0\%), and LaST$_{0}$ (82.0\%). Furthermore, ablating the AWAT mechanism (\textbf{LaST$_{0.5}$ w/o AWAT}) highlights its critical contribution. Although the base model achieves strong results (99.2\% on LIBERO, 85.0\% on RLBench), integrating progress-aware value feedback consistently boosts performance.
These results demonstrate that progress-guided adaptive execution improves task success over the base model without AWAT. Visualizations of the predicted values across task progress are provided in Appendix F.2.

\begin{table}[t]
\centering
\caption{
Comparison on LIBERO, reporting the best success rate (\%) obtained during training.
}
\vspace{-0.2cm}
\label{tab:libero_main}
\resizebox{\columnwidth}{!}{
\begin{tabular}{lccccc}
\toprule
\textbf{Method}
& \textbf{Spatial}
& \textbf{Object}
& \textbf{Goal}
& \textbf{Long}
& \textbf{Mean S.R. $\uparrow$} \\
\midrule
OpenVLA-OFT~\cite{kim2025finetuningvisionlanguageactionmodelsoptimizing}      & 97.6 & 98.4 &  97.9 &  94.5 & 97.1 \\
$\pi_{0.5}$~\cite{intelligence2025pi05visionlanguageactionmodelopenworld} & 98.8 & 98.2 & 98.0 & 92.4 & 96.9 \\
InternVLA~\cite{cai2026internvlaa1unifyingunderstandinggeneration}   & 98.6 & 99.8 & 98.6 & 98.4 & 98.9 \\
Cosmos Policy~\cite{kim2026cosmos} & 98.1 & \textbf{100.0} & 98.2 & 97.6 & 98.5 \\
Motus~\cite{bi2026motus}           & 96.8 & 99.8 & 96.6 & 97.6 & 97.7 \\
Fast-WAM~\cite{yuan2026fast}       & 98.2 & \textbf{100.0} & 97.0 & 95.2 & 97.6 \\
LaST$_{0}$~\cite{liu2026last}          & 99.2 & 99.6 & 98.0 & 95.6 & 98.1 \\
\midrule
LaST$_{0.5}$ w/o AWAT              & 99.6 & \textbf{100.0} & 98.1 & 99.0 & 99.2 \\
\textbf{LaST$_{0.5}$}  & \textbf{99.7} & \textbf{100.0} & \textbf{99.1} & \textbf{99.3} & \textbf{99.5} \\
\bottomrule
\end{tabular}
}
\vspace{-0.2cm}
\end{table}

\begin{figure}[t]
\centering
\includegraphics[width=\linewidth]{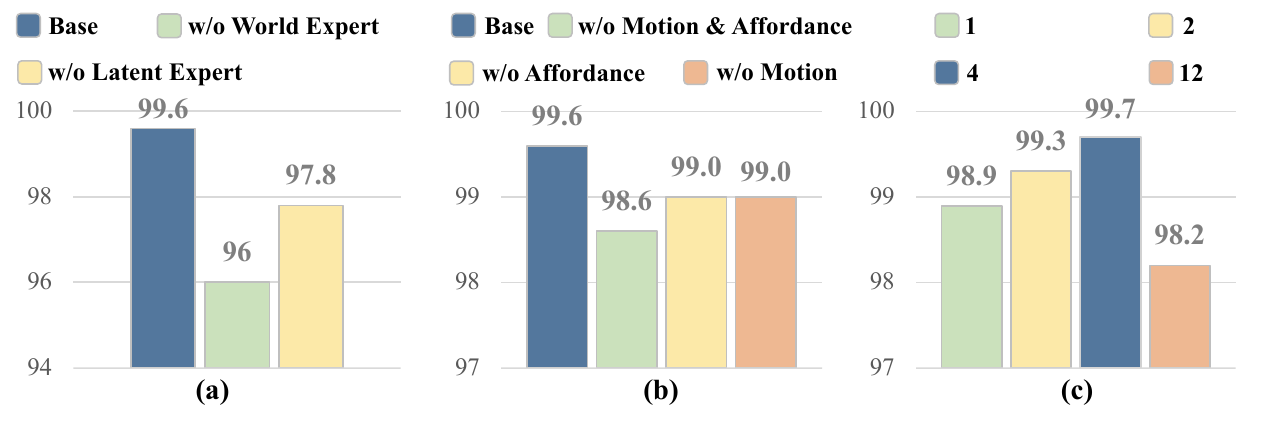}
\vspace{-0.8cm}
\caption{
Ablation studies on (a) individual experts and (b) interaction-centric latent target construction, both using LaST$_{0.5}$ w/o AWAT as the ``Base'' model. (c) Temporal granularity for the value module, evaluated on the full LaST$_{0.5}$.
}
\label{fig:ablation}
\vspace{-0.4cm}
\end{figure}

\begin{figure}[t]
\centering
\includegraphics[width=\linewidth]{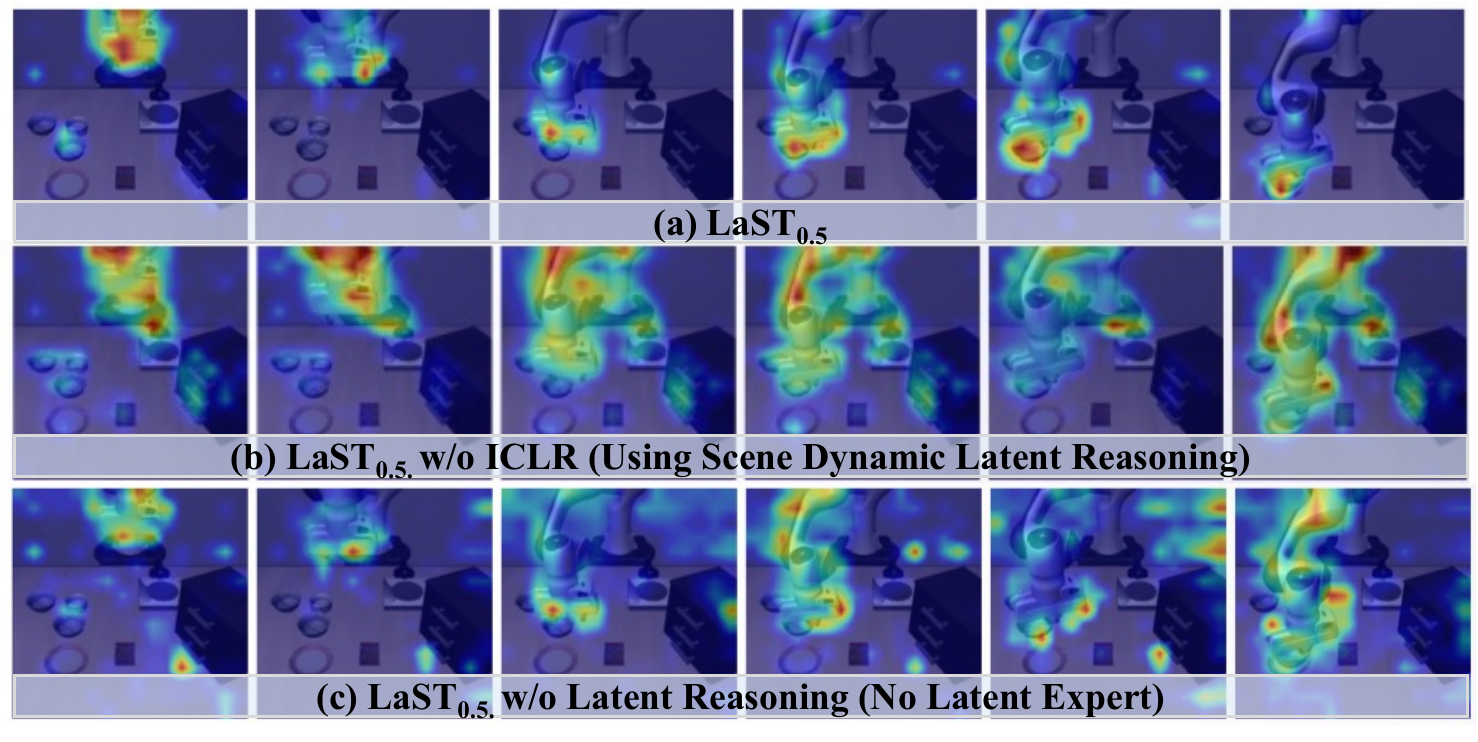}
\vspace{-0.6cm}
\caption{attention maps of LaST$_{0.5}$ and its ablated variants.}
\label{fig:attention_map}
\vspace{-0.4cm}
\end{figure}

\subsection{Ablation Studies}

We conduct ablation studies to evaluate the contribution of
each core component in LaST$_{0.5}$. Unless otherwise specified,
all ablations use LaST$_{0.5}$ without AWAT as the base model,
which achieves 99.6\% on LIBERO-Spatial; enabling AWAT yields
99.7\%. All variants share identical training and evaluation
settings. The temporal-granularity ablation specifically
evaluates the model with AWAT enabled.
Additional ablation results are provided in Appendix E.

\paragraph{Importance of Individual Experts.}
As illustrated in Figure~\ref{fig:ablation} (a), we examine the architectural necessity of each expert by selectively ablating them from our base model (LaST$_{0.5}$ w/o AWAT). Removing the world expert drops the success rate from 99.6\% to 96.0\%, highlighting the importance of temporal and physical dynamics priors. Similarly, removing the latent expert decreases performance to 97.8\%. This confirms its indispensable bridging role in distilling dense global scene features into compact, interaction-centric abstractions for accurate action generation.

\paragraph{Importance of Interaction-Centric Latent Reasoning.}
As shown in Figure~\ref{fig:ablation} (b), we further investigate how specific visual cues within the interaction-centric latent target construction affect action policy learning. Compared to the base model LaST$_{0.5}$ w/o AWAT (99.6\%), we construct three ablated variants:
(1) \textit{w/o Motion \& Affordance}, which uses unmodified future images as latent targets, without motion trajectories or object-box overlays;
(2) \textit{w/o Motion}, which removes end-effector motion curve overlays from latent supervision; and
(3) \textit{w/o Affordance}, which excludes target object affordance cues (e.g., bounding boxes).
These variants exhibit notable performance declines, dropping to success rates of 98.6\%, 99.0\%, and 99.0\%, respectively. This demonstrates that omitting explicit interaction cues causes the policy's attention to diffuse across background distractors, whereas explicitly incorporating spatial trajectory curves and object affordance anchors the latent space to active robot-environment interaction regions, thereby facilitating accurate action generation, as further corroborated by the attention maps visualizations in Figure~\ref{fig:attention_map}.

\paragraph{Temporal Granularity of the Adaptive World-to-Action Transfer.}
As shown in Figure~\ref{fig:ablation}(c), we investigate the
temporal granularity of progress estimation in AWAT. Over a
fixed 12-step prediction horizon, we assign one progress value
to every 1, 2, 4, or 12 action steps, corresponding to 12, 6, 3,
or 1 temporally aligned segments, respectively. On
\textit{LIBERO-Spatial}, the default 4-step granularity achieves
the highest success rate of 99.7\%, compared with 98.9\%,
99.3\%, and 98.2\% for the 1-, 2-, and 12-step settings.
Fine-grained estimation at every step can introduce
high-frequency fluctuations, while a single estimate over all
12 steps provides insufficient temporal resolution for adaptive
truncation. The default setting balances temporal resolution
and stable progress estimation.

\subsection{Real-World Experiment}

\paragraph{Data Collection.}
We evaluate LaST$_{0.5}$ on four challenging real-world manipulation tasks using a dual-arm Tianji Marvin setup, resizing three RGB camera views for the VLM branch (one head view and two wrist views) at a resolution of $224 \times 224$.
Our designed tasks encompass both complex bimanual coordination and fine-grained manipulation:
    1) \textbf{Bag Pulling:} The robot must coordinate both arms to grasp and pull a flexible bag without slipping.
    2) \textbf{Object Organization:} Bimanual pick-and-place to organize cluttered objects into target containers.
    3) \textbf{Screw Fastening:} A high-precision assembly task requiring tight spatial alignment to manipulate a tool and insert a screw into a threaded hole.
    4) \textbf{Parcel-Box Packing:} A long-horizon task requiring the robot to sequentially place objects inside the box, fold the top of the box, and seal the top using a handheld tape dispenser.
Demonstrations are collected using the PICO VR-based teleoperation platform, with $150$ trajectories recorded for each task.
The detailed experimental settings are provided in Appendix C.

\paragraph{Baselines and Evaluation Protocol.}
We compare LaST$_{0.5}$ against three state-of-the-art baselines: $\pi_{0.5}$~\cite{intelligence2025pi05visionlanguageactionmodelopenworld}, Cosmos Policy~\cite{kim2026cosmos}, and LaST$_{0}$~\cite{liu2026last}. For a rigorous and fair evaluation, all policies are trained on the exact same demonstration dataset using identical multi-view visual inputs. Each task is evaluated across $20$ rollouts under varying initial object poses and spatial arrangements. We report full-task success rates for Bag Pulling and Object Organization. For Screw Fastening and Parcel-Box Packing, we report stage-wise success rates. The average success rate is computed across the four tasks based on full-task completion. More details are provided in Appendix D.2.

\begin{table}[t]
\centering
\caption{
Real-world results, reporting success rates (\%) over $20$ rollouts per task. Stage-wise breakdowns are provided for \textit{Screw Fastening} (S1: Tool-use, S2: Insertion) and \textit{Parcel-Box Packing} (S1: Place, S2: Fold, S3: Seal).
}
\vspace{-0.2cm}
\label{tab:real_world}
\resizebox{\columnwidth}{!}{%
\begin{tabular}{lcccccccc}
\toprule
\multirow{2}{*}{\textbf{Method}}
& \multirow{2}{*}{\begin{tabular}[c]{@{}c@{}}
    \textbf{Bag}\\
    \textbf{Pulling}
  \end{tabular}}
& \multirow{2}{*}{\begin{tabular}[c]{@{}c@{}}
    \textbf{Object}\\
    \textbf{Organization}
  \end{tabular}}
& \multicolumn{2}{c}{\textbf{Screw Fastening}}
& \multicolumn{3}{c}{\textbf{Parcel-Box Packing}}
& \multirow{2}{*}{\textbf{Avg. $\uparrow$}} \\
\cmidrule(lr){4-5}
\cmidrule(lr){6-8}
& & & \textbf{S1} & \textbf{S2}
& \textbf{S1} & \textbf{S2} & \textbf{S3} & \\
\midrule
$\pi_{0.5}$
& 75.0 & 85.0
& 80.0 & 30.0
& 95.0 & 75.0 & 20.0
& 52.5 \\
Cosmos Policy
& 60.0 & 85.0
& 60.0 & 20.0
& 60.0 & 35.0 & 10.0
& 43.8 \\
LaST$_{0}$
& 60.0 & 75.0
& 75.0 & 25.0
& 65.0 & 50.0 & 15.0
& 43.8 \\
\midrule
\textbf{LaST$_{0.5}$ (Ours)}
& \textbf{80.0} & \textbf{95.0}
& \textbf{90.0} & \textbf{50.0}
& \textbf{100.0} & \textbf{85.0} & \textbf{30.0}
& \textbf{63.8} \\
\bottomrule
\end{tabular}%
}
\vspace{-0.2cm}
\end{table}

\begin{figure}[t]
\centering
\includegraphics[width=\linewidth]{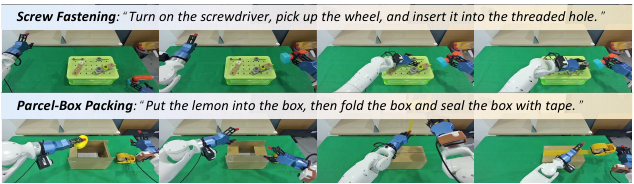}
\vspace{-0.6cm}
\caption{Example visualization of LaST$_{0.5}$ performing real-world tasks, with sequences shown from left to right.}
\label{fig:real_world}
\vspace{-0.4cm}
\end{figure}

\paragraph{Quantitative and Qualitative Results.}
As shown in Table~\ref{tab:real_world}, LaST$_{0.5}$ achieves a leading average success rate of 63.8\% across real-world bimanual tasks, significantly outperforming strong baselines such as $\pi_{0.5}$ (52.5\%) and Cosmos Policy (43.8\%). Beyond robust performance on foundational tasks (e.g., 95.0\% on \textit{Object Organization}), our framework particularly excels in overcoming long-horizon compounding errors. Notably, LaST$_{0.5}$ outperforms $\pi_{0.5}$ in the high-precision insertion stage of \textit{Screw Fastening} (S2: 50.0\% vs. 30.0\%) and the challenging final sealing stage of \textit{Parcel-Box Packing} (S3: 30.0\% vs. 20.0\%). Furthermore, qualitative execution trajectories (Figure~\ref{fig:real_world}) highlight smooth dual-arm synchronization during flexible bag pulling, accurate spatial pose alignment during screw threading, and stable multi-stage execution during parcel packing. Together, these results confirm the real-world robustness of LaST$_{0.5}$ under complex contact dynamics, effectively bridging broad scene understanding with the precise, closed-loop control required for complex physical execution.

\begin{figure}[t]
\centering
\includegraphics[width=\linewidth]{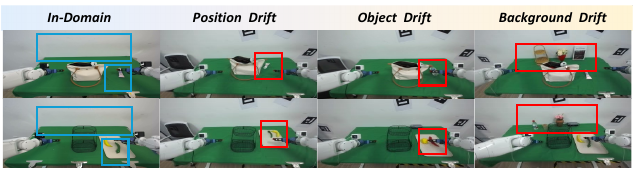}
\vspace{-0.6cm}
\caption{Out-of-distribution scenario setting.}
\label{fig:ood_setting}
\vspace{-0.6cm}
\end{figure}

\paragraph{Generalization Analysis.}
We further evaluate robustness under three out-of-distribution scenarios: variations in object position, object instance, and visual background. As reported in Table~\ref{tab:real_world_generalization}, LaST$_{0.5}$ demonstrates superior adaptability across unseen environmental variations. It shows particular robustness to spatial perturbations (Position shift), achieving a 40.0\% average success rate that more than doubles the performance of strong baselines like $\pi_{0.5}$ (15.0\%) and Cosmos Policy (17.5\%). Additionally, LaST$_{0.5}$ achieves the highest average success under Background shifts (72.5\%)—peaking at an impressive 90.0\% on \textit{Object Organization}—and matches the top baseline performance under Object shifts.
Visualizations are provided in Figure~\ref{fig:ood_setting}.

\begin{table}[t]
\centering
\caption{
Generalization results on two real-world manipulation tasks. We report task-level success rates (\%) under position, object, and background shifts.
}
\vspace{-0.2cm}
\label{tab:real_world_generalization}
\resizebox{\linewidth}{!}{
\begin{tabular}{llccc}
\toprule
\textbf{Method}
& \textbf{Scenario}
& \textbf{\shortstack{Bag Pulling}}
& \textbf{\shortstack{Object Organization}}
& \textbf{Avg. $\uparrow$} \\
\midrule
 $\pi_{0.5}$
& Position   & 15.0 & 15.0 & 15.0 \\
& Object     & \textbf{60.0} & 50.0 & \textbf{55.0} \\
& Background & \textbf{60.0} & 75.0 & 67.5 \\
\midrule
Cosmos Policy
& Position   & 0.0  & 35.0 & 17.5 \\
& Object     & 10.0 & 60.0 & 35.0 \\
& Background & \textbf{60.0} & 65.0 & 62.5 \\
\midrule
LaST$_{0}$
& Position   & 5.0  & 35.0 & 20.0 \\
& Object     & 35.0 & 45.0 & 40.0 \\
& Background & 50.0 & 60.0 & 55.0 \\
\midrule
\textbf{LaST$_{0.5}$ (Ours)}
& Position   & \textbf{25.0} & \textbf{55.0} & \textbf{40.0} \\
& Object     & 45.0 & \textbf{65.0} & \textbf{55.0} \\
& Background & 55.0 & \textbf{90.0} & \textbf{72.5} \\
\bottomrule
\end{tabular}
}
\vspace{-0.2cm}
\end{table}

%% file: Sections/Conclusion.tex
\section{Conclusion}
In this work, we propose LaST$_{0.5}$, a generalist World Action Model that bridges the critical gap between global scene dynamics and localized robotic manipulation.
By introducing Interaction-Centric Latent Reasoning, LaST$_{0.5}$ learns a compact latent space supervised directly by visual cues that capture robot-environment interactions.
To further improve execution robustness, our Adaptive World-to-Action Transfer mechanism dynamically monitors predicted task progress to adapt the execution horizon and trigger future resampling, reducing the impact of stalled or regressing imagined futures on action execution.
Pretrained on over 14k hours of heterogeneous robot data, LaST$_{0.5}$ establishes a new state-of-the-art on the LIBERO and RLBench benchmarks, and demonstrates exceptional long-horizon execution and zero-shot generalization in complex real-world bimanual tasks. Ultimately, LaST$_{0.5}$ presents a highly scalable and reliable paradigm for interaction-centric robotic foundation models.

%% file: Sections/appendix.tex
\setcounter{secnumdepth}{2}

\section{Additional LaST$_{0.5}$ Details}

\subsection{Token Sequence}The foundation of the LaST$_{0.5}$ architecture relies on specific sequence representations across its three branches:
\begin{itemize}
\item \textbf{World Expert:} Models a total of 17 frames, ranging from
$t{-}4$ to $t{+}12$: five historical frames ($t{-}4$ to $t$) serve as clean
observations, while twelve future frames ($t{+}1$ to $t{+}12$) are initialized
from noise. The video frames have a spatial resolution of $256 \times 256$, yielding a $16 \times 16$ spatial token grid (256 tokens) per latent frame. Within the Diffusion Transformer (DiT), the causal video tokenizer
compresses the five historical frames into two 256-token latent frames and the
twelve future frames into three 256-token latent frames, yielding five latent
frames in total. Text information is injected into the World Expert through
cross-attention with these video tokens.
\item \textbf{Latent Expert:} This sequence is formulated purely by concatenating the language instruction tokens and $M$ (default is $3$) latent reasoning query tokens.
\item \textbf{Action Expert:} This sequence is formed by concatenating image tokens, timestamp tokens, and an action chunk consisting of $C$ (default is $24$) continuous action tokens.
\end{itemize}

\subsection{Attention Rules}

\begin{figure}[H]
\centering
\includegraphics[width=\linewidth]{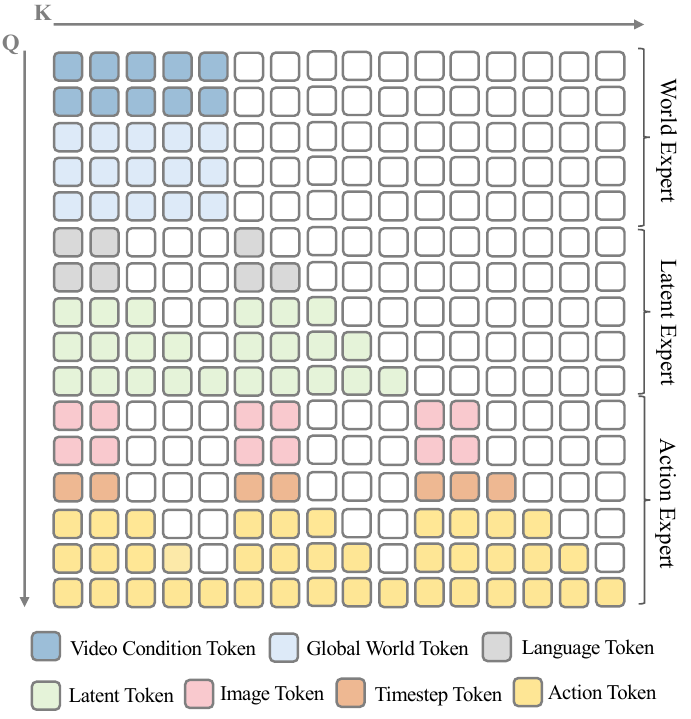}
\caption{Full Attention Mask of the LaST$_{0.5}$ Architecture.
The matrix visualizes the global attention mechanism across the three experts, where rows denote queries ($Q$) and columns denote keys ($K$). Colored cells indicate visible tokens (attention enabled), while white cells denote masked tokens (attention disabled).}
\label{fig:attention_mask}
\end{figure}

The global attention architecture enforces a strict unidirectional information flow: World Expert $\rightarrow$ Latent Expert $\rightarrow$ Action Expert. To synchronize temporal reasoning across modalities, we partition the sequences of predicted future videos, reasoning latents, and continuous action chunks into $M$ temporally aligned segments (where $M=3$ by default). Let $W_i$, $L_i$, and $A_i$ denote the $i$-th segment of the World, Latent, and Action experts, respectively, where $i \in \{1, \dots, M\}$.
As conceptually illustrated by the attention mask in Figure~\ref{fig:attention_mask} (colored cells indicate enabled attention, while white cells denote masking), the visibility rules between and within these segments are defined as follows:

\begin{itemize}
\item \textbf{World Expert:} Operates with global visibility over its own domain but is strictly blind to the Latent and Action Experts. Specifically, any World segment $W_i$ applies full bidirectional attention across all World segments $\{W_1, \dots, W_M\}$ (indicated by the fully colored $M \times M$ blue block).
\item \textbf{Latent Expert:} Follows a step-grained causal masking strategy to restrict direct attention to corresponding and preceding segments. A Latent segment $L_i$ queries information cross-expert from past and current World segments $\{W_1, \dots, W_i\}$, and intra-expert from past and current Latent segments $\{L_1, \dots, L_i\}$. It is masked from seeing any future segments ($W_{j>i}$ or $L_{j>i}$) and remains entirely blind to the Action sequence.
\item \textbf{Action Expert:} Adheres to a similar progressive causal window to ensure action commands are strictly conditioned on corresponding temporal progress. An Action segment $A_i$ can attend cross-expert to World segments $\{W_1, \dots, W_i\}$ and Latent segments $\{L_1, \dots, L_i\}$, as well as intra-expert to Action segments $\{A_1, \dots, A_i\}$. It cannot directly attend to any future segments across any of the three experts.
\end{itemize}
Crucially, this segment-level visibility rule blocks direct attention from earlier latent and action segments to later predictions. The World Expert retains bidirectional attention, consistent with its pretrained architecture, so later world segments can still influence earlier representations indirectly. The mask therefore restricts direct cross-expert conditioning rather than guaranteeing strict temporal isolation.

\subsection{Inference}

The inference execution of LaST$_{0.5}$ processes video predictions, latent reasoning, and action generation through a unified, multi-stage closed-loop pipeline. Initially, the World Expert performs two denoising steps on the noisy future frame tokens ($t+1$ to $t+12$) and stores its Key-Value (KV) cache, while its output visual representations are processed by a Progress-Aware Value Head to yield a task-progress estimate that dynamically determines the number of physical action steps to execute in the current rollout iteration. Subsequently, the latent expert generates all interaction-centric latent tokens in a single parallel forward pass, where a fixed set of learned query tokens produces them simultaneously rather than autoregressively. Finally, the Action Expert performs a single forward pass over its non-action components (image and timestamp tokens) to compute their KV cache, and then executes $n$ (default is 10) iterative denoising steps exclusively over the 24 continuous action tokens representing 12 action steps to synthesize the final action commands.

\section{Large-Scale Pre-training Datasets}
\label{app:pretrain}

\subsection{Dataset Composition}
\label{app:pretrain:composition}

We pre-train LaST$_{0.5}$ on a large-scale, cross-embodiment manipulation
corpus of \textbf{2.48M episodes}, \textbf{536M frames}, and
\textbf{14{,}892 hours}, all standardized to $10$\,Hz
(Table~\ref{tab:pretrain}). The corpus deliberately
spans a spectrum of embodiments and collection modalities: portable
handheld-gripper (\textbf{UMI}) demonstrations, embodied \textbf{dual-arm}
and \textbf{single-arm} teleoperation, \textbf{simulation}, and human
\textbf{egocentric} videos. Portable UMI data forms the majority
($9{,}088$\,h, $61.0\%$), providing scalable, robot-agnostic end-effector
supervision; it is complemented by real-robot teleoperation
(dual-arm $2{,}862$\,h; single-arm $795$\,h), simulation ($1{,}421$\,h),
and human egocentric manipulation ($727$\,h). This composition lets the
world and latent experts absorb broad interaction dynamics from abundant
UMI and human data, while embodied teleoperation grounds the action expert
in real robot control. All heterogeneous action labels are unified into the
relative end-effector action space described in Sec.~B.2.

\begin{table}[t]
\centering
\small
\caption{Composition of the LaST$_{0.5}$ pre-training corpus, sampled at
$10$\,Hz. Sources: AgiBot~\cite{bu2025agibot},
InternData-A1~\cite{tian2025interndata}, DROID~\cite{khazatsky2024droid},
EgoDex~\cite{hoque2025egodex}, RoboCOIN~\cite{wu2025robocoin},
RoboMIND~\cite{wu2024robomind}.}
\setlength{\tabcolsep}{5pt}
\begin{tabular}{llrr}
\toprule
Dataset & Modality & Episodes & Hours \\
\midrule
Genrobot       & UMI          & 656{,}993 & 9{,}088.0 \\
AgiBot         & Dual-arm     & 996{,}381 & 2{,}479.0 \\
InternData-A1  & Sim + Real   & 344{,}607 & 1{,}429.2 \\
DROID          & Single-arm   &  94{,}030 &   766.2 \\
EgoDex         & Human ego    & 314{,}839 &   727.4 \\
RoboCOIN       & Dual-arm     &  42{,}828 &   313.2 \\
RoboMIND       & Dual/Single  &  29{,}763 &    89.6 \\
\midrule
\textbf{Total} & ---          & \textbf{2{,}479{,}441} & \textbf{14{,}892.6} \\
\bottomrule
\end{tabular}
\label{tab:pretrain}
\end{table}

\subsection{Data Alignment and State--Action Standardization}
\label{app:alignment}

Our corpus aggregates heterogeneous sources whose native coordinate systems,
control frequencies, and action conventions differ substantially. To train a single generalist policy over this mixture, we convert every source into one unified state–action standard with an aligned end-effector coordinate system. Following the universal embodied data ingestion paradigm of LDA-1B~\cite{lyu2026lda}, we discard the source-provided action labels and re-derive all states and actions in this common frame.

\paragraph{Unified state.}
Each frame is represented by two keys. The \textbf{wrist} state gives, for each
arm, the 6-DoF end-effector pose $(x,y,z,\text{roll},\text{pitch},\text{yaw})$
expressed in a fixed world frame; we require a right-handed convention so that
the pose is a valid $SE(3)$ transform. When a dataset does not directly expose
the end-effector pose, we recover it by forward kinematics from the robot URDF.
The \textbf{hand} state encodes the end-effector configuration and is handled
separately for the two actuator types: parallel-jaw grippers are normalized to a
scalar in $[-1,1]$ ($-1$ fully closed, $1$ fully open), while dexterous and
human hands follow the fingertip representation in \cite{agarwal2026cosmos},
using five fingertip positions ($5\times3$) expressed in the wrist frame as
$R_{\text{wrist}}^{\top}(t_{\text{fingertip}}-t_{\text{wrist}})$. Gripper and
hand are aligned at the semantic level (forward $=x$, gripper opening direction
$=y$, up $=z$), which lets a single action space span grippers, dexterous hands,
and human hands.

\paragraph{Unified action.}
Rather than reuse the action labels shipped with each source---whose semantics
and quality vary widely---we recompute all actions from the standardized state,
using an actuator-specific rule. For the \textbf{wrist}, we write the pose at
step $t$ as a homogeneous transform
\begin{equation}
T_t=\begin{bmatrix} R_t & t_t \\ \mathbf{0} & 1 \end{bmatrix}\in SE(3),
\end{equation}
where $R_t\in SO(3)$ is the rotation obtained from
$(\text{roll},\text{pitch},\text{yaw})$ and $t_t=(x,y,z)^{\top}$ is the
translation. The action is the \emph{local relative} transform between
consecutive poses, $\delta T = T_t^{-1}T_{t+1}$, which gives
\begin{equation}
\delta R = R_t^{\top}R_{t+1},
\qquad
\delta t = R_t^{\top}(t_{t+1}-t_t),
\end{equation}
so that both rotation and translation are expressed in the local frame and share
consistent semantics across embodiments.
For the \textbf{gripper}, the action encodes the open/close trend: it is set to
open when the state increases beyond a threshold, to close when it decreases
beyond the threshold, and otherwise inherits the previous action; the first
frame is initialized from the measured opening width. For the \textbf{hand}, the
action is simply the next-frame state, i.e. the same $5\times3$ wrist-relative
fingertip representation evaluated at $t{+}1$.

\paragraph{Cross-embodiment alignment.}
Applying the above state and action definitions to heterogeneous sources
requires reconciling their differing coordinate systems and control rates. We
first resample every trajectory to $10$\,Hz and store the per-frame camera
intrinsics and extrinsics for all available views. We then place each trajectory
in a common world frame whose origin is fixed to the left arm's first-frame
position, and calibrate a per-source rotation that maps the native end-effector
axes onto our canonical right-handed convention. For sources whose poses are
expressed in a moving camera frame, we further compose this with the stored
extrinsics to lift the state into the fixed world frame; keeping each
trajectory's state in a fixed frame is what makes the local-relative wrist
action well defined. With states thus aligned and the gripper/hand
representations unified as described above, the wrist, gripper, and hand actions
are then recomputed directly from the standardized state.

\section{Real-world Setup}

\begin{figure}[t]
\centering
\includegraphics[width=\linewidth]{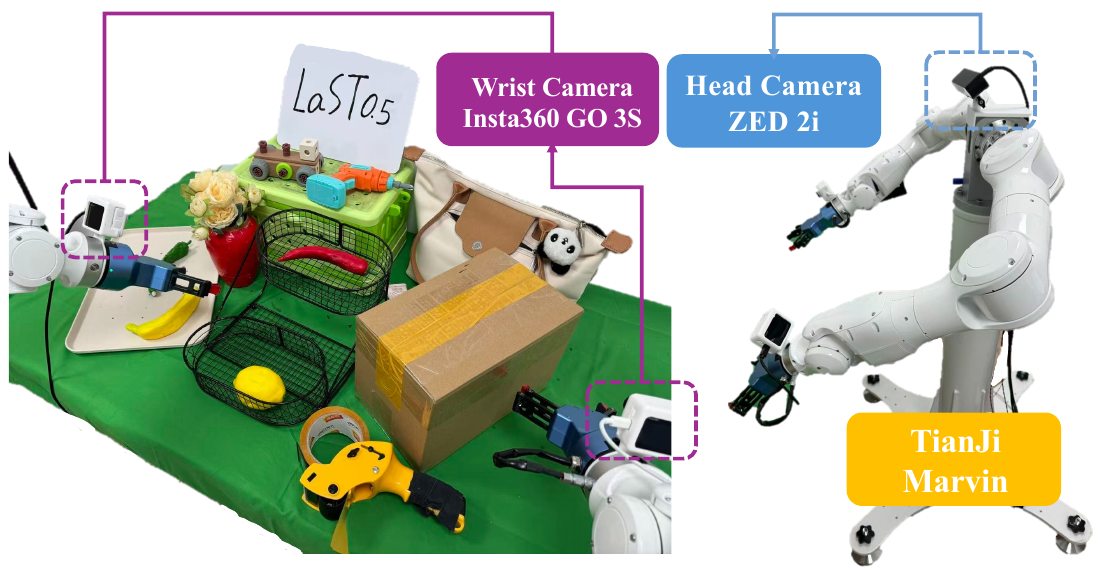}
\caption{Real-World Robot Experiment Setup.}
\label{fig:setup}
\end{figure}

The physical deployment of LaST$_{0.5}$ relies on a 7-DoF Tianji Marvin parallel-gripper setup.
Across all tasks, the observation suite for the VLM branch consists of three RGB views resized to $224 \times 224$: one head-mounted ZED 2i camera providing an egocentric scene view, and two wrist-mounted Insta360 GO 3S cameras capturing close-range embodiment-object interactions around the end effectors.

Robot demonstrations are collected using Pico controller teleoperation to provide end-effector commands together with gripper open/close commands. All teleoperated trajectories are synchronized with the three camera streams and stored as language-conditioned observation-action sequences for downstream training and evaluation.
All hardware assets utilized in our real-world setup are illustrated in Figure~\ref{fig:setup}.

\section{Self Collected Data and Success-Judgement Protocol}

\subsection{Self Collected Data}
We collect 150 demonstrations for each of the four real-world manipulation tasks, which cover bimanual coordination, deformable-object manipulation, fine-grained tool use, and long-horizon execution.

\paragraph{Bag Pulling.}
The robot first uses its right gripper to pick up a tabletop object and place it inside a backpack. It then grasps the right side of the backpack with the right gripper to stabilize the bag, while the left gripper grasps the zipper pull and draws it across the opening to close the backpack.

\paragraph{Object Organization.}
The left gripper stabilizes a two-layer storage basket while the right gripper picks up fruits and vegetables from the tabletop and sorts them into separate layers of the basket, placing fruits in the lower layer and vegetables in the upper layer.

\paragraph{Screw Fastening.}
The robot performs a high-precision assembly task across two sequential stages. \textit{S1}: Tool-use, where the right gripper grasps the lower part of the control switch on an electric screwdriver to activate rotation in the fastening direction; grasping the upper part would instead reverse the rotation direction. \textit{S2}: Insertion, where the left arm picks up the wheel and inserts it into the threaded hole on the vehicle body.

\paragraph{Parcel-Box Packing.}
The robot performs a long-horizon packing task that requires coordinated bimanual manipulation across three sequential stages. \textit{S1}: Place, the left arm picks up an object and places it inside the parcel box; \textit{S2}: Fold, both arms fold and press the left, right, front, and rear flaps in sequence; \textit{S3}: Seal, the right arm grasps the tape dispenser, the left arm pulls out the tape, and the right arm attaches it to the left side of the box. The left arm then stabilizes the box and presses the tape firmly against it while the right arm pulls the tape across the box and tears it off.

\subsection{Success-Judgement Protocol}
For \textit{Bag Pulling} and \textit{Object Organization}, we report the full-task success rate, and a rollout is counted as successful only when the entire task is completed. For \textit{Screw Fastening} and \textit{Parcel-Box Packing}, we evaluate the completion of the original stages (S1--S2 and S1--S3, respectively). A stage is considered successful only when it is fully completed; partial completion or a stage not reached is counted as unsuccessful. Each stage's success rate is the percentage of successful stages across all 20 evaluation rollouts.

\paragraph{Bag Pulling.}
A trial is considered successful only if the object is placed completely inside the backpack and the zipper is fully closed without the robot losing its grasp on the bag or the zipper.

\paragraph{Object Organization.}
A trial is considered successful only if all objects are placed completely inside the basket and every object is sorted into its designated layer.

\paragraph{Screw Fastening.}
\textit{S1} is considered successful only if the robot grasps the electric screwdriver and activates rotation in the correct direction. \textit{S2} is considered successful only if the robot picks up the wheel and completes insertion into the threaded hole on the vehicle body. Full-task success requires completing both stages and is measured by the S2 success rate.

\paragraph{Parcel-Box Packing.}
\textit{S1} is considered successful only if the robot picks up the object and places it inside the box. \textit{S2} is considered successful only if the robot folds all four flaps and presses them flat. \textit{S3} is considered successful only if the robot picks up the tape dispenser, pulls out the tape, attaches it to the side of the box, and completes the tape sealing. Full-task success requires completing all three stages and is measured by the S3 success rate.

\section{Additional Ablation Study}
\label{app:ablation}

We report three additional ablations of the interaction-centric
latent reasoning module on LIBERO-Spatial. All variants disable
AWAT and use matched training and evaluation settings, differing
only in the design choice under study. The baseline is
LaST$_{0.5}$ without AWAT (99.6\%), using three latent tokens
($t{+}4,t{+}8,t{+}12$), a frozen SigLIP encoder for latent targets,
and parallel learned-query generation.

\subsection{Latent CoT Capacity}
\label{app:ablation:capacity}
We evaluate the required capacity of the latent CoT by varying the number of latent tokens---controlled by the temporal sampling density of future frames. As Table~\ref{tab:abl_capacity} demonstrates, maintaining 3 latent tokens yields the strongest success rate (99.6). Conversely, expanding the token budget to 6 or 12 provides no additional gain and causes a slight decline in performance.
This indicates that a sequence of 3 latent tokens is already sufficient to effectively represent the critical interaction dynamics of the future scene. Encoding denser frame sequences does not provide stronger representational power. Instead, it introduces informational redundancy that can complicate optimization and lead to slight overfitting to local inter-frame variations. Given that additional tokens increase computational overhead without benefiting action selection, we adopt 3 latent tokens as our optimal setting.

\begin{table}[H]
\centering\small
\caption{Latent CoT capacity. We ablate the number of latent tokens by varying the temporal sampling density.}
\setlength{\tabcolsep}{6pt}
\begin{tabular}{lcc}
\toprule
Latent tokens & Sampling & Success rate \\
\midrule
3 (base) & $t{+}4,8,12$      & 99.6 \\
6        & $t{+}2,4,\dots,12$ & 99.4 \\
12       & $t{+}1,2,\dots,12$   & 99.3 \\
\bottomrule
\end{tabular}
\label{tab:abl_capacity}
\end{table}

\subsection{Latent Target Source}
\label{app:ablation:gtsource}
We investigate different strategies for constructing the latent targets by comparing the action expert's native visual encoder (SigLIP) against various feature representations extracted from a video generation model (Cosmos). As shown in Table 7, while all variants achieve highly competitive results ($\geq 99.1$), extracting latent targets using the base SigLIP encoder yields the highest success rate ($99.6$). We hypothesize that this is primarily due to representation alignment. Utilizing the action expert's native visual encoder ensures that the generated latent targets are projected into the exact semantic space that the downstream policy is already optimized to decode. Conversely, leveraging external representations---even those from powerful world models or post-DiT layers---introduces a slight domain gap between the planning space and the action generation space. Therefore, using the native encoder provides the most direct and effective conditioning signal for the policy.

\begin{table}[H]
\centering\small
\caption{Latent target source. Performance comparison of different representations for latent targets.}
\setlength{\tabcolsep}{6pt}
\begin{tabular}{lc}
\toprule
Latent target source & Success rate \\
\midrule
SigLIP (base)                         & 99.6 \\
Cosmos, pre-DiT                       & 99.1 \\
Cosmos, post-DiT                      & 99.4 \\
Cosmos (action-conditioned), post-DiT & 99.2 \\
\bottomrule
\end{tabular}

\label{tab:abl_gtsource}
\end{table}
\subsection{Latent Generation Mechanism}
\label{app:ablation:generation}
We evaluate different mechanisms for generating latent reasoning tokens during training: parallel generation via learned query tokens in a single forward pass, standard autoregressive decoding (predicting tokens sequentially before action generation), and open-loop teacher forcing (conditioning on ground-truth latents). As presented in Table~\ref{tab:abl_generation}, the parallel and autoregressive schemes achieve comparable, near-optimal performance at convergence ($99.6$ vs.\ $99.4$). Given that the parallel approach synthesizes all latent tokens simultaneously, it delivers autoregressive-level reasoning quality while drastically reducing inference latency. Consequently, we adopt parallel generation as our default strategy. In contrast, teacher forcing exhibits a noticeable performance degradation, dropping to $97.5$. We attribute this $2.1$-point decline to exposure bias---a well-known train--test mismatch. When trained exclusively on perfect ground-truth latents, the action expert becomes overly reliant on them and struggles to recover from the distribution shift introduced by its own generated, potentially imperfect latents during inference. This performance gap highlights the necessity of conditioning on model-generated latents during the training phase to ensure robust closed-loop deployment.

\begin{table}[H]
\centering\small
\caption{Latent generation mechanism.}
\setlength{\tabcolsep}{6pt}
\begin{tabular}{lc}
\toprule
Generation & Success rate \\
\midrule
Parallel (base) & 99.6 \\
Autoregressive  & 99.4 \\
Teacher forcing & 97.5 \\
\bottomrule
\end{tabular}
\label{tab:abl_generation}
\end{table}

\section{Additional Visualization}
\subsection{Qualitative Analysis of Action-to-Vision Attention}
\label{app:visualization}
\begin{figure}[h]
\centering
\includegraphics[width=\linewidth]{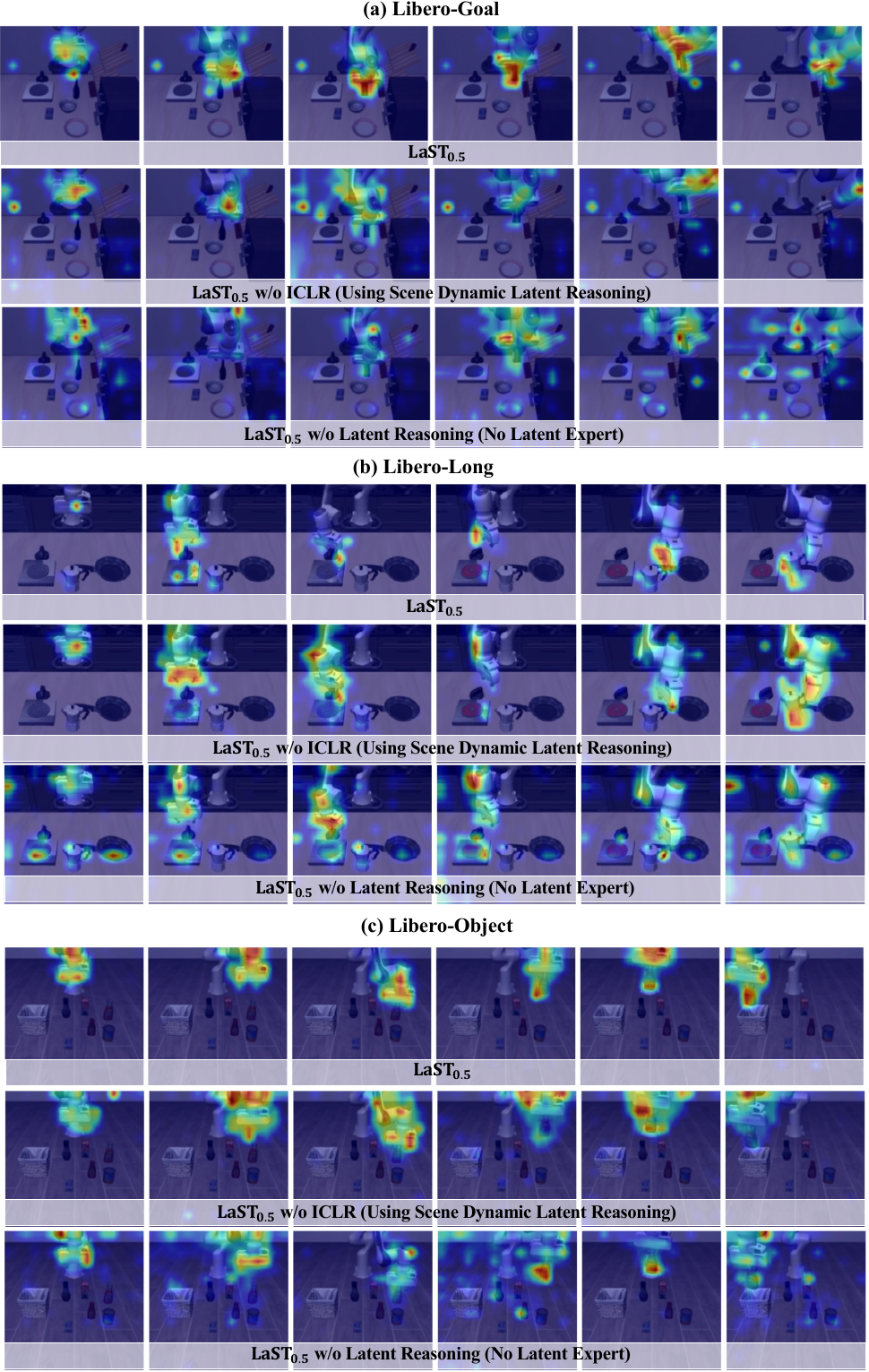}
\caption{Action-to-vision attention maps.}
\label{fig:appx_attnmap}
\end{figure}

To further understand the impact of our latent reasoning mechanism, we visualize the action-to-vision cross-attention maps across the LIBERO-Goal, LIBERO-Long, and LIBERO-Object suites (Figure~\ref{fig:appx_attnmap}). These visualizations complement the LIBERO-Spatial results presented in the main paper.
As shown in the top row, our default method (LaST$_{0.5}$) exhibits highly concentrated attention on the end-effector and the specific target objects being manipulated. This sharp spatial focus shows that the model prioritizes regions where robot–object interactions occur. In contrast, when the latent expert is forced to model the global \textit{scene dynamics} (middle row), the attention becomes noticeably dispersed. By attempting to reconstruct or predict the entire future scene, the model wastes representational capacity on irrelevant background elements or static objects, diluting the focus on the actual manipulation bottleneck. Finally, the baseline without any latent reasoning (bottom row) fails entirely to isolate the regions of interest, resulting in scattered, blurry, and temporally lagging attention maps. This stark contrast qualitatively validates that reasoning specifically about interaction dynamics provides qualitative evidence of improved interaction-focused grounding, complementing the policy-level ablations.

\subsection{Analysis of Value Progress Trajectories}
\label{sec:visualization_value}
\begin{figure}[h]
\centering
\includegraphics[width=\linewidth]{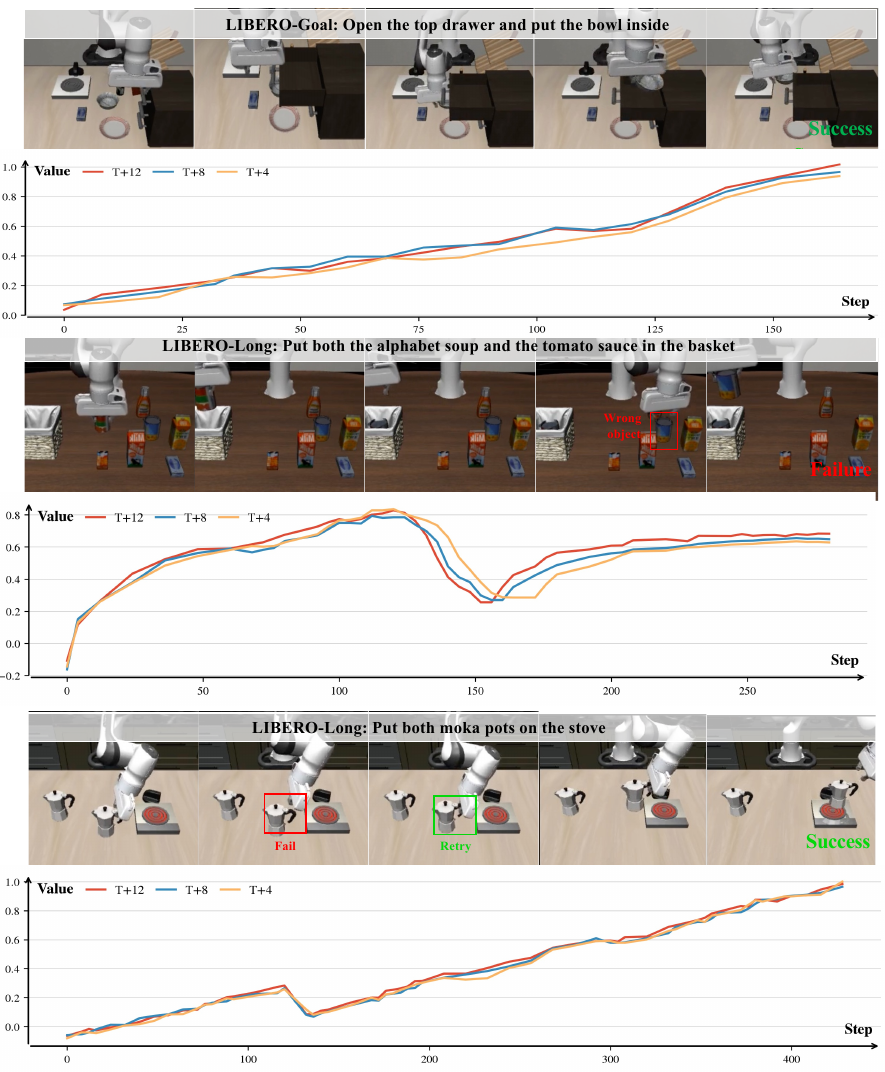}
\caption{Value progress visualization. The value curve acts as a real-time indicator of task status. It illustrates predicted task progress, rising during successful execution and dropping in the shown failure cases involving misaligned world knowledge.}
\label{fig:appx_value}
\end{figure}

Figure~\ref{fig:appx_value} illustrates the temporal evolution of the estimated value during task execution. Derived from the predicted video features, this value acts as a reliable proxy for prediction accuracy and task progress. The trajectories reveal a consistent pattern: a rising value strongly correlates with successful, well-aligned action execution (as seen in the top row and the recovery phase of the bottom row). Conversely, a sudden drop in the value curve serves as an accurate early warning of execution failure. As demonstrated in the middle row, such drops typically occur when the action generation is negatively influenced by misaligned world knowledge, causing the policy to deviate from the correct target. This confirms the efficacy of our value head as an accurate, real-time indicator of execution robustness.

\subsection{Additional Real-World Task Executions}
\label{sec:Additional Real-World Task Executions}
\begin{figure}[h]
\centering
\includegraphics[width=\linewidth]{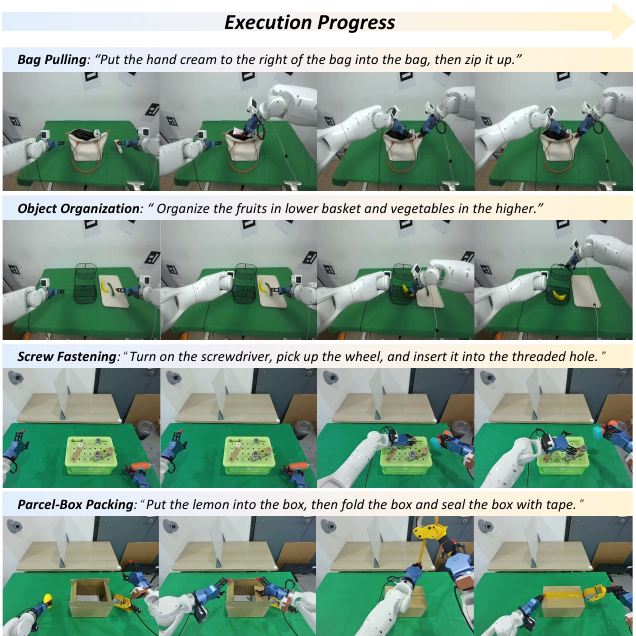}
\caption{Additional visualization of LaST$_{0.5}$ performing real-world tasks.}
\label{fig:real_world_apdx}
\end{figure}

Figure~\ref{fig:real_world_apdx} visualizes LaST$_{0.5}$ executing additional real-world tasks. The sequences illustrate coordinated bimanual manipulation in Bag Pulling and Parcel-Box Packing, semantic and spatial grounding in Object Organization, and precise tool use in Screw Fastening. These examples demonstrate how the policy translates multi-stage language instructions into coordinated action sequences across different manipulation settings, complementing the quantitative evaluation of task success and generalization in the main paper.

\section{Failure Case Analysis}
\label{sec:failure_analysis}
\begin{figure}[H]
\centering
\includegraphics[width=\linewidth]{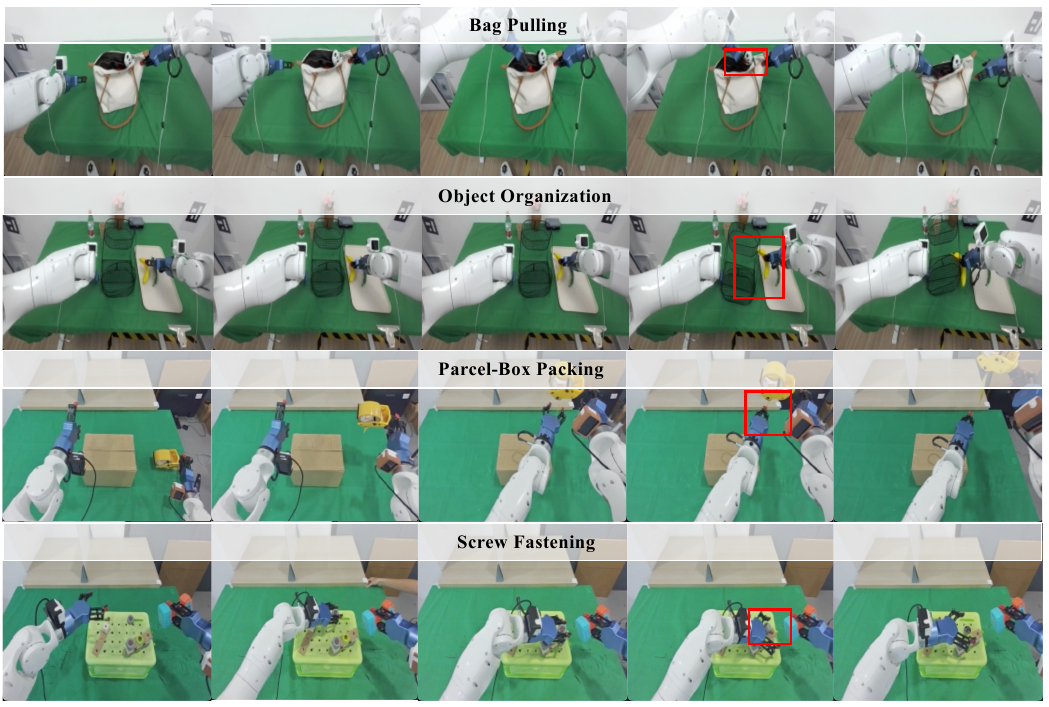}
\caption{Illustrative failure sequences from four real-world tasks. Red bounding boxes indicate the exact instant and location where each execution error occurs.}
\label{fig:real_world_fail}
\end{figure}

Figure~\ref{fig:real_world_fail} illustrates typical execution failures in our real-world experiments, highlighting the physical boundaries of the current system. In the \textbf{Bag Pulling} task, the execution halts due to unexpected mechanical resistance when the zipper gets stuck. The \textbf{Object Organization} failure occurs when the grasped cucumber collides with and shifts the basket, indicating slight inaccuracies in modeling spatial geometries. In \textbf{Parcel-Box Packing}, the policy executes the correct pulling motion, but the tape slips due to insufficient grip force and friction, revealing a hardware limitation of the parallel-jaw gripper. Finally, the \textbf{Screw Fastening} task fails due to a millimeter-level misalignment between the screw and the hole, underscoring the challenge of completing tight-tolerance insertions relying solely on visual feedback.

